\documentclass[smallextended]{svjour3}

\smartqed  
\usepackage{booktabs}
\usepackage{enumitem}
\usepackage{longtable}
\usepackage{graphicx}
\usepackage{array}
\usepackage{dcolumn}
\usepackage{amsfonts}
\usepackage{amssymb}
\usepackage{xspace}
\usepackage{xfrac}
\usepackage{array}
\usepackage{multirow}
\usepackage{xcolor}
\usepackage{footnote}
\usepackage{url}
\usepackage{makeidx}
\usepackage{epsfig}
\usepackage{amsmath}
\usepackage{multicol}
\usepackage{multirow}
\usepackage{mathtools}
\usepackage[misc]{ifsym}
\usepackage[ruled,linesnumbered]{algorithm2e}
\usepackage{tabularx}
\usepackage{array}
\usepackage{footnote}
\makesavenoteenv{tabular}
\makesavenoteenv{table}

\begin{document}

\title{Model Selection for Time Series Forecasting}
\subtitle{An Empirical Analysis of Multiple Estimators}

\titlerunning{Model Selection for Time Series Forecasting}

\author{Vitor~Cerqueira,
Luis~Torgo, and Carlos~Soares}

\authorrunning{V. Cerqueira et al.}

\institute{Vitor Cerqueira (\Letter) \at
         Dalhousie University, Halifax, Canada\\
         \email{vitor.cerqueira@dal.ca}
         \and
         Luis Torgo \at
         Dalhousie University, Halifax, Canada\\
         \email{ltorgo@dal.ca}
         \and
         Carlos Soares \at
         Fraunhofer AICOS Portugal, Porto, Portugal\\
         INESC TEC, Porto, Portugal\\
         University of Porto, Porto, Portugal\\
         \email{csoares@fe.up.pt}
         }

\date{Received: date / Accepted: date}

\maketitle

\begin{abstract}

Evaluating predictive models is a crucial task in predictive analytics. 
This process is especially challenging with time series data where the observations show temporal dependencies. 
Several studies have analysed how different performance estimation methods compare with each other for approximating the true loss incurred by a given forecasting model. 
However, these studies do not address how the estimators behave for model selection: the ability to select the best solution among a set of alternatives. 
In this paper, we address this issue and compare a set of estimation methods for model selection in time series forecasting tasks. We attempt to answer two main questions: (i) how often is the best possible model selected by the estimators; and (ii) what is the performance loss when it does not. 
Using a case study that comprises 3111 time series we found that the accuracy of the estimators for selecting the best solution is low, despite being significantly better relative to random selection. Moreover, the overall forecasting performance loss associated with the model selection process ranges from 0.28\% and 0.58\%. We also discovered that the sample size of time series is an important factor in the relative performance of the estimators.

\end{abstract}

\section{Introduction}\label{intro}

Estimating the predictive performance of models using the available data is a crucial stage in the data science pipeline. 
We study this problem for time series forecasting tasks, in which the time dependency among observations poses a challenge to several estimation methods that assume observations are independent.

\begin{figure}[h]
    \centering
    \includegraphics[width=\textwidth, trim=0cm 0cm 0cm 0cm, clip=TRUE]{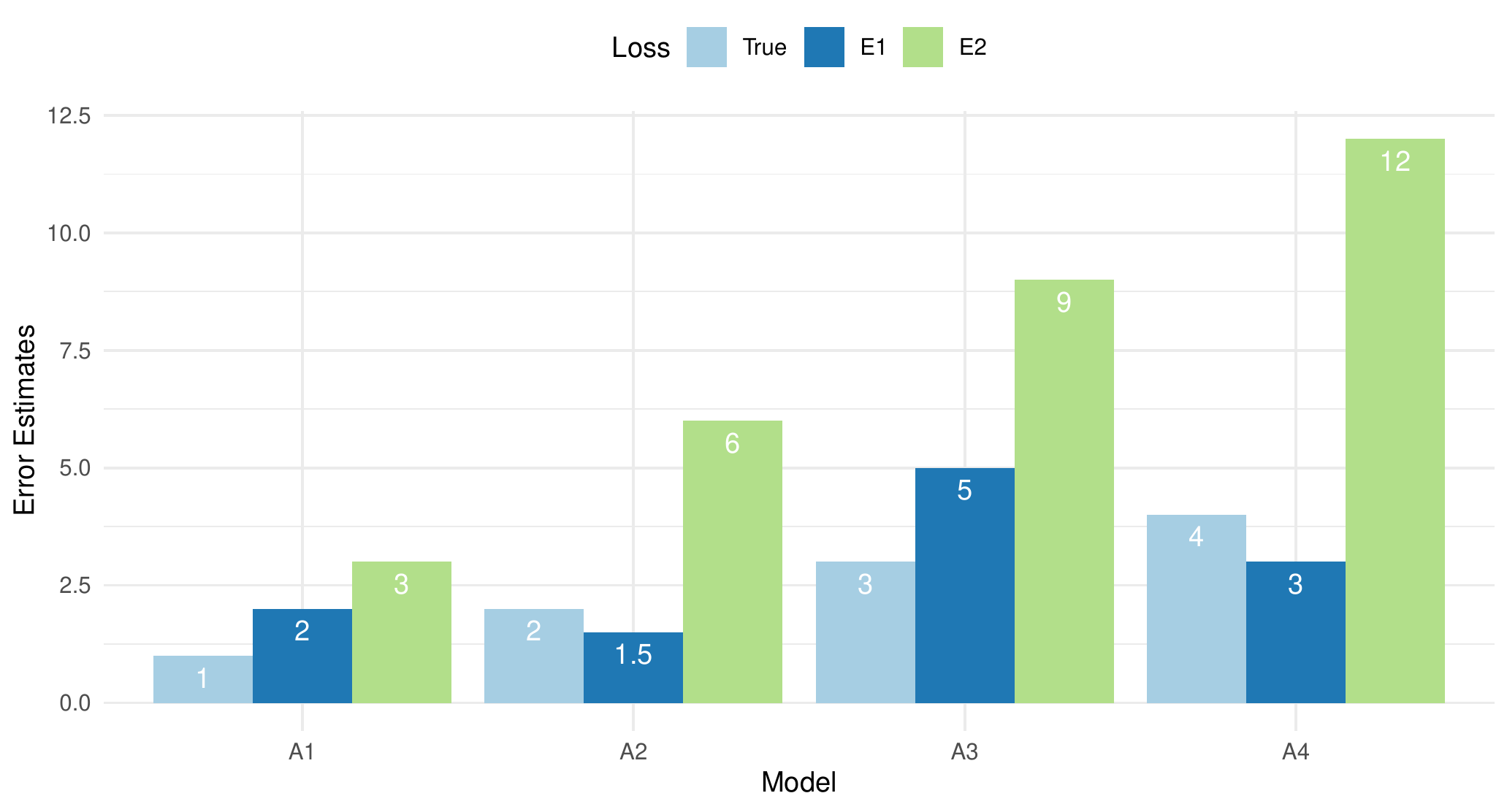}
    \caption{True error (labeled True) and two error estimates (using estimators E1 and E2) of the predictive performance of four models: A1, A2, A3, and A4. The estimator E1 shows, on average, the best estimates. However, E2 produces a perfect ranking of the models contrary to E1.}
    \label{fig:example}
\end{figure}

Performance estimation methods are used to solve two main tasks: (i) to provide a reliable estimate of performance in order to inform the end-user of the expected generalization ability of a given predictive model; and (ii) to use these estimates to perform model selection, i.e., to select a predictive model among a set of possible alternatives that can actually be different parameter settings of the same model. A substantial amount of work addresses the first of these tasks for time series forecasting problems, e.g.~\cite{bergmeir2012use,bergmeir2018note,cerqueira2020evaluating,tashman2000out,mozetivc2018evaluate}. In this work, we address the second problem. Although similar, model selection and performance estimation are two different problems \cite{breiman1992submodel,arlot2010survey}. On the same data set, an estimator may provide the best loss estimations, on average, but not the best model rankings for selection purposes. This idea is illustrated in Figure~\ref{fig:example}.

In this example, there are four predictive models: A1, A2, A3, A4, which are shown in the x-axis of Figure \ref{fig:example}. The true test set loss is depicted by the light blue bars. Thus, the correct ranking of the models is A1 $>$ A2 $>$ A3 $>$ A4. A1 is the best model has it shows the lowest test loss. 
Two estimators, E1 and E2, are used to approximate the error of each model.
The estimator E1 produces the best loss approximations (nearest to the true error), on average, relative to estimator E2. However, its estimated ranking (A2 $>$ A1 $>$ A4 $>$ A3) is actually different than the true ranking and worse than the one produced by E2. Despite providing worse performance estimates, E2 outputs a perfect ranking of the models. This example shows that one estimator is better for performance estimation (E1), and the other for model selection (E2). This matter was explored before by Breiman and Spector \cite{breiman1992submodel} for i.i.d. data sets. They found interesting differences in the relative performance of estimation methods when applied to model selection and performance estimation.

Our goal in this paper is to study the ability of different estimators (e.g. K-fold cross-validation) for model selection in time series forecasting tasks, in which the observations are not i.i.d..
Given a pool of alternative models, we study: (i) how often the best solution is picked (the one which maximizes forecasting performance on test data); and (ii) how much performance is lost when it does not. 
In other words, we analyse the ability of different estimators to rank the available predictive models by their performance in unseen observations. We are particularly interested in the top ranked model, which is the one most probably selected for deployment, i.e. for predicting future observations of the domain under study.

Most estimation procedures will involve repeating the application of a model to different test sets/folds. The estimation results across these different folds are typically combined using the arithmetic mean. The selected model among a set of considered alternatives is the one with lowest average estimated error. In this paper, we study the possibility of combining the results across folds using the average rank, instead of the average estimated error. The average rank is a non-parametric approach which may be beneficial to smooth the effect of large errors (outliers) in particular folds \cite{yang2007consistency}.

We carried a set of experiments using 10 estimation methods, 3111 time series data sets, and 50 auto-regressive forecasting models. The results show that the accuracy of the estimators for selecting the most appropriate solution ranges from 7\% to 10\%. Overall, the forecasting performance loss incurred due to incorrect the model selection  ranges from 0.28\% to 0.58\%. 
We also controlled the experiments by the sample size of time series and found interesting differences in the relative performance. Particularly, the performance loss is considerably larger for smaller time series. Finally, regarding the strategy for combining the results across folds, we found that the average rank leads to a comparable performance relative to the average error.

In summary, the contribution of this paper is an extensive study comparing a set of performance estimation methods for model selection in time series forecasting tasks. To our knowledge, this paper is the first to quantify the impact of using a particular estimator for model selection in forecasting problems. The experiments carried out in this paper are available online\footnote{\url{https://github.com/vcerqueira/model_selection_forecasting}}.

This paper is organised as follows. In the next section we review the work related to this paper. In Section \ref{sec:problem_definition}, we formalise the predictive task relative to time series forecasting and the model selection process. The experiments are presented in Section \ref{sec:experiments} and discussed in Section \ref{sec:discussion}. Finally, the paper is concluded in Section \ref{sec:conclusions}.

\section{Related Work}\label{sec:related_work}

In this section, we provide an overview of the literature related to our work. We briefly describe performance estimation methods designed for time series forecasting models (Section \ref{sec:rw_performance_estimation}). We also list previous works which study these methods and highlight our contributions. In Section \ref{sec:ml_for_forecasting}, we overview the literature on the application of machine learning methods for forecasting.

\subsection{Performance Estimation in Forecasting}\label{sec:rw_performance_estimation}

Several approaches for evaluating the predictive performance of models in time series have been proposed in the literature. We discuss terminology before describing these. According to Arlot and Celisse \cite{arlot2010survey}, cross-validation denotes the process of splitting the data, once or multiple times, for estimating the performance of a predictive model. In this process, part of the data is used to fit a model while the remaining observations are used for testing. All methods we analyse in this work follow this procedure. However, we refer to cross-validation as the class of approaches that follow the splitting criteria of K-fold cross-validation and which assume independence among data points. Other procedures, for example prequential \cite{dawid1984present}, work under different assumptions. Therefore, we designate them accordingly.

Arguably, the most common approach to assess the performance of a forecasting model is a holdout procedure, also known as out-of-sample evaluation (e.g. \cite{bergmeir2018note}). The initial part of the time series is used to fit the model, and the last part of the data is used to test it. Tashman \cite{tashman2000out} recommends applying this approach in multiple testing periods. Indeed, Cerqueira et al. \cite{cerqueira2020evaluating} show that holdout applied in multiple, randomized, periods leads to the best estimation ability relative to several other state of the art estimators for non-stationary time series.

K-fold cross-validation works by randomly assigning the available observations to K different equally-sized folds. Each fold is then iteratively used for testing a predictive model which is built using the remaining observations.
This process is theoretically inadequate to evaluate time series models because the observations are not independent \cite{arlot2010survey,opsomer2001nonparametric}.
However, Bergmeir et al. \cite{bergmeir2012use,bergmeir2018note} show that cross-validation approaches can be successfully applied to estimate the performance of forecasting models. For example, they showed that, in some scenarios, the K-fold cross-validation procedure described above provides better results than out-of-sample evaluation \cite{bergmeir2018note}.
Notwithstanding, several methods have been proposed as extensions to the K-fold cross-validation, which aim at mitigating the problem of the dependence among observations. These include the blocked cross-validation \cite{snijders1988cross}, modified cross-validation \cite{mcquarrie1998regression}, and \textit{hv}-blocked cross-validation \cite{racine2000consistent}.
In different works, both Bergmeir et al. \cite{bergmeir2012use} and  Cerqueira et al. \cite{cerqueira2020evaluating} compare different estimators for evaluating forecasting models. They suggest using the blocked form of K-fold cross-validation for stationary time series.

The prequential method is also a common evaluation approach for time-dependent data \cite{dawid1984present}. This method is also referred to as time series cross-validation or time series split by practitioners. Prequential denotes the process in which an observation (or batch of observations) is first used for testing a predictive model and then to update it, and it is the most common solution in data stream mining tasks \cite{gama2009evaluating}. For more general time series, prequential approaches are typically applied in contiguous blocks of data. 
Moreover, prequential can be applied in different manners. For example, using a growing window or a sliding window. 

The above-mentioned estimation methods are described in more detail in Section \ref{sec:algorithms}.
These have been studied in different works according to how well they approximate the loss that a predictive model incurs in a test set \cite{bergmeir2012use,bergmeir2018note,cerqueira2020evaluating}.
The objective was to assess: (i) the magnitude of their estimation errors, and (ii) the direction of the error, i.e., whether the estimators under-estimate or over-estimates the loss of the respective model. 
These two quantities allow us to analyse which estimators provide the most reliable approximations, on average. They enable the quantification of the generalization ability of models, which help the end-user decide whether or not a model can be deployed.
However, as illustrated in Figure \ref{fig:example}, the estimator with the best approximations is not necessarily the one presenting the best ranking ability, and thus, the most appropriate for model selection \cite{breiman1992submodel}. Correctly identifying the relative performance of predictive models is an important feature for model selection. Contrary to related works, we focus on analysing estimation methods from this perspective.

\subsection{Machine Learning Methods for Forecasting}\label{sec:ml_for_forecasting}

Without loss of generality, and as we will describe in Section \ref{sec:algorithms}, we focus on typical machine learning regression algorithms for forecasting. We apply these in an auto-regressive manner using time-delay embedding, which is formalized in Section \ref{sec:problem_definition}. In this section, we overview several works which address the applicability of machine learning approaches to forecasting.

Makridakis et al. \cite{makridakis2018statistical} reported that machine learning approaches performed worse relative to traditional methods such as ARIMA \cite{chatfield2000time}, exponential smoothing \cite{gardner1985exponential}, or a simple seasonal random walk. These conclusions were drawn from 1045 monthly time series with low sample size (an average of 118 observations). However, in a more recent work, Makridakis and his colleagues \cite{spiliotis2020comparison} show that machine learning approaches provide better forecasting performance for SKU demand forecasting when compared to these traditional methods. Moreover, Cerqueira et al. \cite{cerqueira2019machine} show that the conclusions drawn in \cite{makridakis2018statistical} are only valid for small time series.  
In the M5 forecasting competition \cite{makridakis2020m5}, the \texttt{lightgbm} gradient boosting method \cite{ke2017lightgbm}, which is a popular machine learning algorithm, was the method used in the winning solution and some of the runner ups.\footnote{\url{https://github.com/Mcompetitions/M5-methods}}

Other studies have also shown that standard regression methods can be successfully applied to time series forecasting problems. Cerqueira et al. \cite{cerqueira2017dynamic,cerqueira2019arbitrage} develop a dynamic ensemble for forecasting. The ensemble is heterogeneous and comprised by several machine learning methods, along with other traditional forecasting approaches. Corani et al. \cite{corani2020automatic} propose a method based on Gaussian processes for automatic forecasting. Deep neural networks are increasingly applied to this type of problems and several architectures have been developed, such as N-BEATS \cite{oreshkin2019n} or DeepAR \cite{salinas2020deepar}, among others. Another notable work is that by Smyl \cite{smyl2020hybrid}, which developed a hybrid method which combines a recurrent neural network with exponential smoothing.

\section{Problem Definition}\label{sec:problem_definition}

In this section, we define two tasks. First, we formalise the time series forecasting problem from an auto-regressive perspective (Section \ref{sec:autoregression}). Then, we define the model selection problem (Section \ref{sec:rw_model_selection}). Finally, in Section \ref{sec:avgrank} we overview the average rank method for algorithm selection across multiple data sets and how we apply it for model selection within a single data set.

\subsection{Auto-regression}\label{sec:autoregression}

Let $Y$ denote a time series $Y = \{y_1, y_2, \dots, y_n\}$, in which $y_i$ is the $i$-th out  of $n$ time-ordered observations. The goal is to predict the future values of this time series. Without loss of generality, in this work we focus on one-step ahead forecasting. This means that we predict the next value of the time series based on its historical observations.

The predictive task can be formalized using time delay embedding by constructing a set of observations of the form ($X$, $y$). 
The target value $y_i$ is modelled based on the past $p$ values before it: $X_i = \{y_{i-1}, y_{i-2}, \dots, y_{i-p} \}$. This process leads to a multiple regression problem where each $y_i \in \mathcal{Y} \subset \mathbb{R}$ represents the \textit{i}-th observation we want to predict, and $X_i \in \mathcal{X} \subset \mathbb{R}^p$ represents the respective explanatory variables.
Effectively, the time series is transformed into the data set $\mathcal{D}(X,y) = \{X_i, y_i\}^{n}_{p+1}$.

\subsection{Model Selection}\label{sec:rw_model_selection}

\begin{figure}[h]
    \centering
    \includegraphics[width=.8\textwidth, trim=0cm 0cm 0cm 0cm, clip=TRUE]{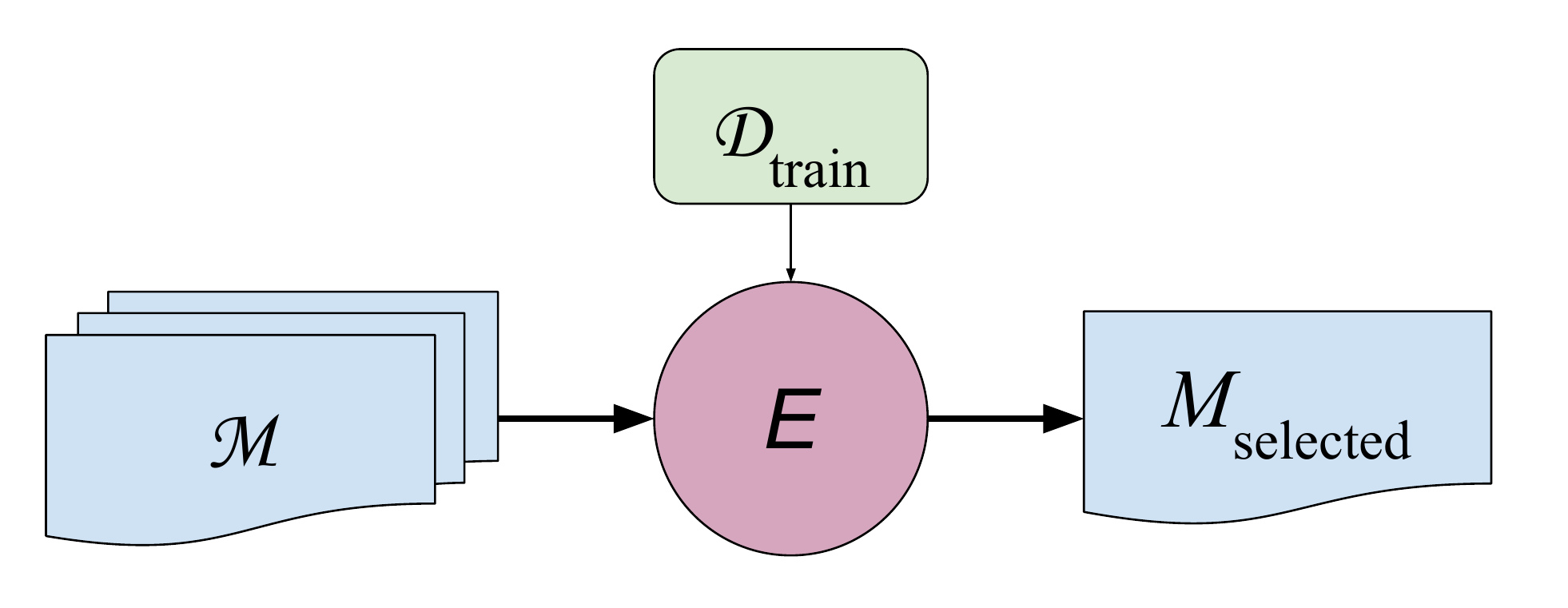}
    \caption{The typical workflow for model selection. The estimation method E tests different algorithms $\text{A} \in \mathcal{A}$ using the training data. The algorithm with best estimations is selected for predicting future observations.}
    \label{fig:scheme}
\end{figure}

Model selection denotes the process of using the available training data to select a predictive model M among a set of \textit{m} alternatives $\mathcal{A} = \{A_1, \dots, A_m\}$. As depicted in Figure~\ref{fig:scheme}, each alternative is evaluated using an estimation method E, such as holdout repeated in multiple testing periods, and a training data set $\mathcal{D}_{train}$. The model which maximizes the predictive performance according to the estimator is selected and used in future observations (a test set).

The goal is to find and select $\text{A}_* \in \mathcal{A}$, which is the model with the best performance on the test set. The model selection problem can then be formalized as follows:

\begin{equation}\label{eq:1}
    \text{A}_{\text{selected}} \in \underset{\text{A} \in \mathcal{A}}{\text{argmin}} \big( L ( \text{A}, \mathcal{D}_{train}, E ) \big)
\end{equation}

\noindent where L represents the loss metric that quantifies the expected error of a predictive models, and E is the estimation method applied to estimate such measure.

In Equation~\ref{eq:1}, the function $L$ takes as input the learning algorithm, the training data, and the estimation method. The output of $L$ is the expected error of the respective learning algorithm. Then, the model $\text{A}_{\text{selected}}$ is selected, which is the one that minimizes the expected error. Ideally, $\text{A}_{\text{selected}}$ represents $\text{A}_*$, which denotes the model with best performance in a test set. However, this is not necessarily the case as $E$ is only able to provide an approximation of the true error of any model. Our goal in this paper is to analyse a set of estimation methods $\mathcal{E} = \{E_1, E_2, \dots, E_q\}$ according to their ability to find $\text{A}_*$, and their behavior when they do not (specifically, assess how much performance is lost). This can be regarded as an analysis of the ranking ability of the estimators, $\mathcal{E}$, in which we are particularly interested in the top ranked model. 
In  Section \ref{sec:experiments} we present a set of experiments that compare a set of estimators from this perspective.

\subsection{Average Rank}\label{sec:avgrank}

The average rank is a procedure which is typically applied to carry out a statistical comparison of multiple learning algorithms over multiple data sets \cite{brazdil2000comparison}. 
The rank of a predictive model denotes its position in terms of performance relative to its competitors. A rank of 1 in a given data set means that the respective model was the best performing one. Effectively, the average rank represents the average relative position of a given predictive model.
As Benavoli et al. \cite{benavoli2016should} explain, the average rank is a non-parametric approach which does not assume normality of the sample means and is robust to outliers. This approach is often used for algorithm selection \cite{abdulrahman2018speeding}.

In this work, we hypothesize that the average rank may be a useful approach for model selection within a single data set. Following Equation \ref{eq:1}, the selected model is typically the one which minimizes the expected error. This expected error is estimated by averaging (with the arithmetic mean) the error of each model across a number of folds. However, the average error is amenable to outliers. For example, a model may incur into a large error in a single fold which will significantly effect its average. On the other hand, the average rank approach may amplify small errors \cite{brazdil2000comparison}.
In this context, we test whether the average rank is a better approach to combine the results of multiple models across multiple folds within the same problem.

Yang \cite{yang2007consistency} explored this idea before in the context of regression tasks. The author refers to this process as voting cross-validation, and concluded that it leads to a comparable performance relative to the standard error-based cross-validation. In this work, we attempt to make this comparison for time series forecasting problems.

\section{Experiments}\label{sec:experiments}

The experiments carried out in this paper aim at comparing different estimators for model selection in time series forecasting tasks.
They are designed to address the following research questions:

\begin{enumerate}
    \item \textbf{RQ1}: What is the accuracy of different performance estimation methods for model selection? That is, how often do their estimates lead to picking the best model (the one which maximizes the true predictive performance in new observations);
    \item \textbf{RQ2}: What is the forecasting performance lost when performance estimation methods do not pick the best model?
    \item \textbf{RQ3}: Do the experimental results vary when controlling for the sample size of time series?
    \item \textbf{RQ4}: How does the average rank (voting cross-validation) compare with the average error for aggregating the results across folds for model selection?
    \item \textbf{RQ5}: What is the relative execution time of each estimator?
\end{enumerate}

\subsection{Data Sets}

The experiments in this paper are carried out using 3111 real-world time series. 
We retrieved all the daily time series with at least 500 observations from the M4 case study \cite{makridakis2020m4}. Our query included the sample size condition (at least 500 data points) as it is an important component for the training of machine learning models \cite{cerqueira2019machine}. This query returned 2937 time series. These time series are from several domains of application, including demographics, finance, industry, macro-economics, and micro-economics. 
The remaining 174 time series were retrieved from a previous related study \cite{cerqueira2020evaluating}.  149 out of these 174 time series were retrieved from the benchmark database \textit{tsdl} \cite{tsdlpackage}. We also included 25 time series used by Cerqueira et al. \cite{cerqueira2019arbitrage}. 

\subsection{Experimental Design}

We create a realistic scenario to compare the different performance estimation methods for model selection. We start by splitting the available time series into two parts: an estimation set, which contains the initial 70\% of observations; and a test set, which contains the subsequent 30\% observations. The general goal is to select the model which provides the best performance on the test set. In practice, we cannot perform direct estimations on this set as it represents future observations. Therefore, we resort to the estimation set, which represents the available data, to estimate which is the best model.

For each estimator $E_i \in \mathcal{E}$, we carry out the model selection process illustrated in Figure \ref{fig:scheme} and defined in Equation \ref{eq:1}, in which the training data represents the estimation set.

This process results in a set of models $\{\text{A}_{E_1}, \dots, \text{A}_{E_q}\}$, where $A_{E_i}$ is the model selected by estimator $E_i$. 
Finally, we evaluate each estimator according to its model selection ability using the test set. The evaluation process is described in the next section.

\subsection{Evaluation}

Let $\text{A}_E$ denote the model selected by the estimation method E, and $\text{RMSE}(\text{A}_E)$ its generalization root mean squared error in a given time series problem. Hopefully, $\text{A}_E$ is equal to $\text{A}_*$, which represents the best model that should be selected. In such case, $\text{RMSE}(\text{A}_E)$ would be optimal according to the pool of available models. 

We use $\text{RMSE}(\text{A}_E)$ to quantify and compare different estimation methods. We compute the percentage difference between the error of the model selected by the estimator E ($\text{RMSE}(\text{A}_E)$) and the error of $\text{A}_*$. This can be formalized as follows:

\begin{equation}\label{eq:2}
    \text{LOSS}(E) = \frac{\text{RMSE}(\text{A}_E) - \text{RMSE}(\text{A}_*)}{\text{RMSE}(\text{A}_*)} \times 100
\end{equation}

\noindent where $\text{LOSS}(E)$ denotes the generalization error associated with the estimator $E$ for a given data set. This error is zero in the case E selects $\text{A}_*$, which is the correct choice. Otherwise, there is a positive performance loss associated with the model selection process, which is quantified by Equation \ref{eq:2}.

If we carry this analysis with multiple time series problems, from the definition in Equation \ref{eq:2} we can compute the following statistics to summarise the quality of a performance estimation method for the task of model selection:
\begin{itemize}
    \item Accuracy: how often a performance estimation method picks the best possible forecasting model. This can be quantified as the proportion of times that $\text{RMSE}(E)$ is equal to zero;
    
    \item Average Loss (AL): The average loss incurred by picking the wrong forecasting model (not selecting $\text{A}_*$). This loss is quantified by taking the average of each $\text{RMSE}(E)$ across all the problems when $\text{A}_E$ is different than $\text{A}_*$;
    
    \item Overall Average Loss (OAL): A combination of the two previous measurements: we compute AL but also taking into account when E selects $\text{A}_*$, in which case the loss is zero.

\end{itemize}

The three metrics listed above allow us to compare different estimators according to their ability to select the best forecasting model among a set of alternatives.

\subsection{Learning Algorithms and Estimation Methods}\label{sec:algorithms}

For each time series problem, each estimator compares 50 alternative models, and selects the one which maximizes the expected performance. The models are obtained using different parameter settings of the following learning algorithms: support vector regression \cite{karatzoglou2004kernlab}, multivariate adaptive regression splines \cite{earth}, random forests \cite{ranger2015}, projection pursuit regression \cite{friedman1981projection}, rule-based regression based on Cubist \cite{Cubist2014}, multi-layer perceptron \cite{monmlp2017}, generalized linear regression \cite{glmnet2010}, Gaussian processes \cite{karatzoglou2004kernlab}, principal components regression \cite{plspackage}, partial least squares regression \cite{plspackage}, and extreme gradient boosting \cite{chen2016xgboost}. The algorithms and respective parameters are described in Table~\ref{tab:expertsspecs}.

\begin{table}[!thb]
	\centering
	\caption{Summary of the learning algorithms}		
	\begin{tabular}{llll}
	\toprule
	\textbf{ID} & \textbf{Algorithm} & \textbf{Parameter} & \textbf{Value}\\
	\midrule
	    \multirow{4}{*}{\texttt{SVR}} & \multirow{4}{*}{Support Vector Regr.} & \multirow{2}{*}{Kernel} & \{Linear, RBF\\
	    
	    & & & Polynomial, Laplace\}\\
	    
	    & & Cost & \{1, 5\}\\
	    
	    & & $\epsilon$ & \{0.1, 0.01\}\\
	    
	    \midrule   
	    
	    \multirow{3}{*}{\texttt{MARS}} & \multirow{3}{*}{Multivar. A. R. Splines} & Degree & \{1, 3\} \\
	    
	    & & No. terms & \{5, 10, 20\} \\
	    
	    & & Forward thresh. & \{0.001\} \\
	    
	    \midrule   
	    
	    \multirow{2}{*}{\texttt{RF}} & \multirow{2}{*}{Random forest} & No. trees & \{250, 500\} \\
	    
	    & & Mtry & \{5, 10\} \\
	    
        \midrule   
        
        \multirow{2}{*}{\texttt{PPR}} & \multirow{2}{*}{Proj. pursuit regr.} & No. terms & \{2, 4\} \\
	    
	    & & Method & \{super smoother, spline\} \\
	    
	    \midrule   
        
        \texttt{RBR} & Rule-based regr. & No. iterations & \{1, 5, 10, 25\}\\
        
        \midrule   
	    
	    \multirow{2}{*}{\texttt{MLP}} & \multirow{2}{*}{Multi-layer Perceptron} & Units Hid. Lay. 1 & \{10, 15\} \\
	    
	    & & Units Hid. Lay. 2 & \{0, 5\}\\
	    
	    \midrule   
	    
        \texttt{GLM} & Generalised Linear Regr. & Penalty mixing & \{0, 0.25, 0.5, 0.75, 1\}\\
        
        \midrule   
        
        \multirow{3}{*}{\texttt{GP}} & \multirow{3}{*}{Gaussian Processes} & \multirow{2}{*}{Kernel} & \{Linear, RBF,\\

	    & & & Polynomial, Laplace\}\\
	    
	    & & Tolerance & \{0.001\}\\
	    
	    \midrule   
        
        \texttt{PCR} & Principal Comp. Regr. &  \textit{Default} & - \\
        
        \midrule   
        
        \texttt{PLS} & Partial Least Regr. & Method & \{kernel, SIMPLS\} \\
        
        \midrule   
        
        \texttt{XGB} & Gradient Boosting & \textit{Auto}\footnote{Automatically optimized using a grid search based on the R package \textit{tsensembler}} & - \\

		\bottomrule    
	\end{tabular}%
	\label{tab:expertsspecs}
\end{table}

We focus on regression learning algorithms, which have been shown competitive forecasting performance relative to traditional approaches such as ARIMA \cite{chatfield2000time} or exponential smoothing \cite{gardner1985exponential} (c.f. Section \ref{sec:ml_for_forecasting}).

In the experiments we apply a total of 10 performance estimation methods, which are described below:

\begin{itemize}
    \item \texttt{K-fold cross-validation} (\texttt{CV}): First, the time series observations are randomly shuffled and split into K folds. Then, each fold is iteratively selected for testing. A model is trained on K-1 folds, and tested in the remaining one. This approach breaks the temporal order of observations, which is problematic for dependent data such as time series \cite{arlot2010survey}. However, it has been shown that \texttt{CV} is applicable in some time series scenarios \cite{bergmeir2018note};
    \item \texttt{Blocked K-fold cross-validation} (\texttt{CV-Bl}): This approach is identical to \texttt{CV}. The difference is that \texttt{CV-Bl} does not shuffle the observations before assigning them to different folds. This leads to K folds of contiguous observations. Bergmeir et al \cite{bergmeir2012use} and Cerqueira et al \cite{cerqueira2020evaluating} recommend this approach for estimating the performance of forecasting models if the time series is stationary;
    
    \item \texttt{Modified Cross-validation} (\texttt{CV-Mod}): The modified cross-validation is a variant of \texttt{CV} which attempts at decreasing the dependency between training and testing observations \cite{mcquarrie1998regression}. First, time series observations are randomly shuffled into K folds, similarly to \texttt{CV}. Then, in each iteration of the cross-validation procedure, some of the training observations are removed. Particularly, we remove the training observations which are within $p$ (the size of the auto-regressive process) observations of any testing point. While this process increases the independence among observations, a considerable number of observations are removed;
    
    \item \texttt{hv-Blocked K-fold cross-validation} (\texttt{CV-hvBl}): The hv-blocked K-fold cross-validation \cite{racine2000consistent} is a variant of \texttt{CV-Bl}. Similarly to \texttt{CV-Mod}, it removes some instances to decrease the dependency among observations. Specifically, for each iteration, the adjacent $p$ observations between training and testing are removed. This process creates a small gap between the two sets;

    \item \texttt{Holdout}: This method represents the typical out-of-sample estimation approach, in which the final part of the time series is held out for testing. This process runs in a single iteration, in which the initial 70\% of observations are used for training, while the subsequent 30\% ones used for testing;
    
    \item \texttt{Repeated Holdout} (\texttt{Rep-Holdout}): An extension of \texttt{Holdout} in which this process is repeated K times in multiple, randomized, testing periods \cite{tashman2000out,cerqueira2020evaluating}. 
    For each one of the K iterations, a random point is chosen in the time series. Then, the 60\% observations (out of the total time series length \textit{n}) before this point are used for training, while the subsequent 10\% observations (out of \textit{n}) are used for testing. Note that the window for selecting this random point is restricted by the size of the training and testing sets \cite{cerqueira2020evaluating};

    \item \texttt{Prequential in Blocks} (\texttt{Preq-Bls}): We apply the prequential evaluation methodology using K blocks of data and a growing window \cite{dawid1984present}. 
    In the first iteration, the initial block containing the first \textit{n}/K observations is used for training, while the subsequent block (also containing \textit{n}/K observations) is used for testing. Afterwards, these two blocks are merged together, and used for training in the second iteration. In this iteration, the third block of data is used for testing. 
    This process continues until the last block is tested. The \texttt{Preq-Bls} procedure is commonly used for evaluating time series models. This method is often referred to as time series cross-validation\footnote{\url{https://robjhyndman.com/hyndsight/tscv/}}\textsuperscript{,}\footnote{The \textit{model\_selection} module from the \textit{scikit-learn} Python library designates this method as \textbf{TimeSeriesSplits}};
    
    \item \texttt{Prequential in Sliding Blocks} (\texttt{Preq-Sld-Bls}): This method represents a variant of \texttt{Preq-Bls} but applied with a sliding window. This means that, after each iteration, the oldest block of data is discarded. Therefore, in each iteration a single block of observations is used for training, and another one is used for testing;
    
    \item \texttt{Trimmed Prequential in Blocks} (\texttt{Preq-Bls-Trim}): A variant of \texttt{Preq-Bls} in which the initial splits are discarded due to low sample size: The initial iterations use a training sample size that may not be representative of the complete available time series, which may bias the results. Formally, according to this method, only the final 60\% of the K iterations of the \texttt{Preq-Bls} procedure are considered. For example, if \texttt{Preq-Bls} splits the time series into 10 blocks, the method \texttt{Preq-Bls-Trim} only considers (and averages) the results on the last 60\% (i.e. 6) iterations;
    
    \item \texttt{Prequential in Blocks with a Gap} (\texttt{Preq-Bls-Gap}): A final variant of \texttt{Preq-Bls}, in which a gap is introduced between the training and testing sets. In each iteration, there is a block of \textit{n}/K observations splitting the training and test sets. Similarly to \texttt{CV-Mod} e \texttt{CV-hvBl}, the motivation for this process is to increase the independence between the two sets.
\end{itemize}

\noindent We refer to the work by Cerqueira et al. \cite{cerqueira2020evaluating} for more information on these methods, including visual representations. 
We use two approaches to combine the results of the K iterations  of each estimator: the average error according to the arithmetic mean, and the average rank. 

\subsection{Parameter Settings}

The number of lags \textit{p} (c.f. Section~\ref{sec:autoregression}) used in each time series is set according to the False Nearest Neighbors method \cite{kennel1992determining}, which is a common approach in non-linear time series analysis. The number of folds or repetitions \textit{K} of the estimation methods is set to 10, where applicable. 

\subsection{Results}

We present the main results in Table \ref{tab:results_general}, which shows the score of each estimator over the 3111 time series, both for the average error and average rank approaches. In the table, the methods are ordered by increasing values of OAL, obtained either by average rank or average error. 

\subsubsection{RQ1: Accuracy of the Estimators}

\begin{table}[ht]
\centering
\caption{Results for all method over the 3111 time series. Methods are ordered by increasing values of overall average loss. Each metrics is bold according to which approach (average rank or average error) provides the best result for each estimator.}
\label{tab:results_general}
\begin{tabular}{lrrr | rrr}
  \hline \\[0.005cm]
  
 & \multicolumn{3}{c}{\textbf{Average Error}} & \multicolumn{3}{c}{\textbf{Average Rank}}\\[0.1cm]
 \hline 
 & \textbf{Acc.} & \textbf{AL} & \textbf{OAL} &  \textbf{Acc.} & \textbf{AL} &  \textbf{OAL} \\ 
  \hline
\texttt{CV-Bl} & \textbf{0.10} & 0.34$\pm$0.72 & 0.29$\pm$0.68 & 0.08 & \textbf{0.32$\pm$0.64} & \textbf{0.28$\pm$0.64} \\ 

  \texttt{Preq-Bls} & \textbf{0.09} & \textbf{0.33$\pm$0.70} & \textbf{0.28$\pm$0.68} & 0.08 & 0.34$\pm$0.68 & 0.30$\pm$0.66 \\ 
  
  \texttt{CV-hvBl} & \textbf{0.10} & 0.35$\pm$0.72 & 0.30$\pm$0.69 & 0.09 & \textbf{0.33$\pm$0.64} & \textbf{0.28$\pm$0.63} \\ 
  
  \texttt{CV-Mod} & \textbf{0.07} & \textbf{0.33$\pm$0.68} & \textbf{0.30$\pm$0.66} & 0.06 & 0.35$\pm$0.70 & 0.31$\pm$0.68 \\ 
  
  \texttt{Preq-Bls-Gap} & \textbf{0.09} & \textbf{0.36$\pm$0.79} & \textbf{0.30$\pm$0.75} & 0.07 & 0.37$\pm$0.73 & 0.32$\pm$0.70 \\
  
  \texttt{CV} & \textbf{0.09} & \textbf{0.38$\pm$0.82} & \textbf{0.32$\pm$0.78} & 0.07 & 0.40$\pm$0.74 & 0.35$\pm$0.73 \\ 
  
  \texttt{Preq-Sld-Bls} & \textbf{0.07} & 0.39$\pm$1.06 & 0.34$\pm$1.03 & 0.06 & \textbf{0.37$\pm$0.99} & \textbf{0.33$\pm$0.93} \\
  
  \texttt{Preq-Bls-Trim} & \textbf{0.09} & 0.41$\pm$0.86 & 0.35$\pm$0.83 & 0.08 & \textbf{0.40$\pm$0.76} & \textbf{0.35$\pm$0.74} \\
  
  \texttt{Rep-Holdout} & \textbf{0.08} & \textbf{0.42$\pm$0.88} & \textbf{0.35$\pm$0.85} & 0.07 & 0.40$\pm$0.79 & 0.36$\pm$0.75 \\ 
  
  \texttt{Holdout} & 0.07 & 0.67$\pm$1.79 & 0.58$\pm$1.64 & 0.07 & 0.67$\pm$1.79 & 0.58$\pm$1.64 \\ 
   \hline
\end{tabular}
\end{table}

The accuracy of the estimators for finding the best predictive model ranges from 7\% to 10\%.
These values represent a significant improvement relatively to a random selection procedure, which has an expected accuracy of 2\% (1 over 50 possible alternative models).
Notwithstanding, this degree of accuracy across all methods means that a given estimator will most probably fail to select the most appropriate model in the available pool. In relative terms, \texttt{CV-Bl} and \texttt{CV-hvBl} show the best score (10\%) while \texttt{CV-Mod}, \texttt{Preq-Sld-Bls}, and \texttt{Holdout} show the worst one (7\%). 

Overall, the accuracy scores for finding the best predictive model highlights the importance of studying the subsequent shortfall: how much forecasting performance is lost by picking the wrong model.

\subsubsection{RQ2: Performance Loss During Model Selection}

The AL and OAL scores quantify the average performance loss of the estimator. We present both for completeness but will focus on the OAL metric in our analysis because it also incorporates the cases in which the estimator selects the correct model.   

The scores of OAL range from 0.28\% (\texttt{CV-Bl}, \texttt{Preq-Bls}, and \texttt{CV-hvBl}) and 0.58\% (\texttt{Holdout}). As explained before, this metric represents the average (median) difference between the loss of the forecasting model selected by the respective estimator and the loss of the best possible solution available, in which the average is computed across all time series.
Essentially, if we apply one of the best estimators we can expect a performance loss of about 0.28\% with respect to the best predictive model in the available pool. Note that these results are highly dependent on the particular pool of available forecasting algorithms.
The values for average loss are significant. In many domains of application, each increment in forecasting performance has a considerable financial impact within organizations \cite{jain2017answers}. Therefore, it is crucial to maximize this performance. However, unless the domain is sensitive to small differences in performance, the chosen estimation method is not a critical factor for performance.

Overall, except for \texttt{Holdout} which shows an OAL significantly higher than the rest, the results are comparable across all estimation methods. Especially when considering the dispersion (IQR), which is quite significant.
Even the standard cross-validation procedure (\texttt{CV}), which is a poor method for performance estimation for forecasting \cite{cerqueira2020evaluating}, only shows a OAL difference of 0.04\% to the best estimators. This shows that breaking the temporal order of observations during model selection is, in general, not problematic for time series forecasting tasks. 

\begin{figure}[h]
    \centering
    \includegraphics[width=\textwidth, trim=0cm 0cm 0cm 0cm, clip=TRUE]{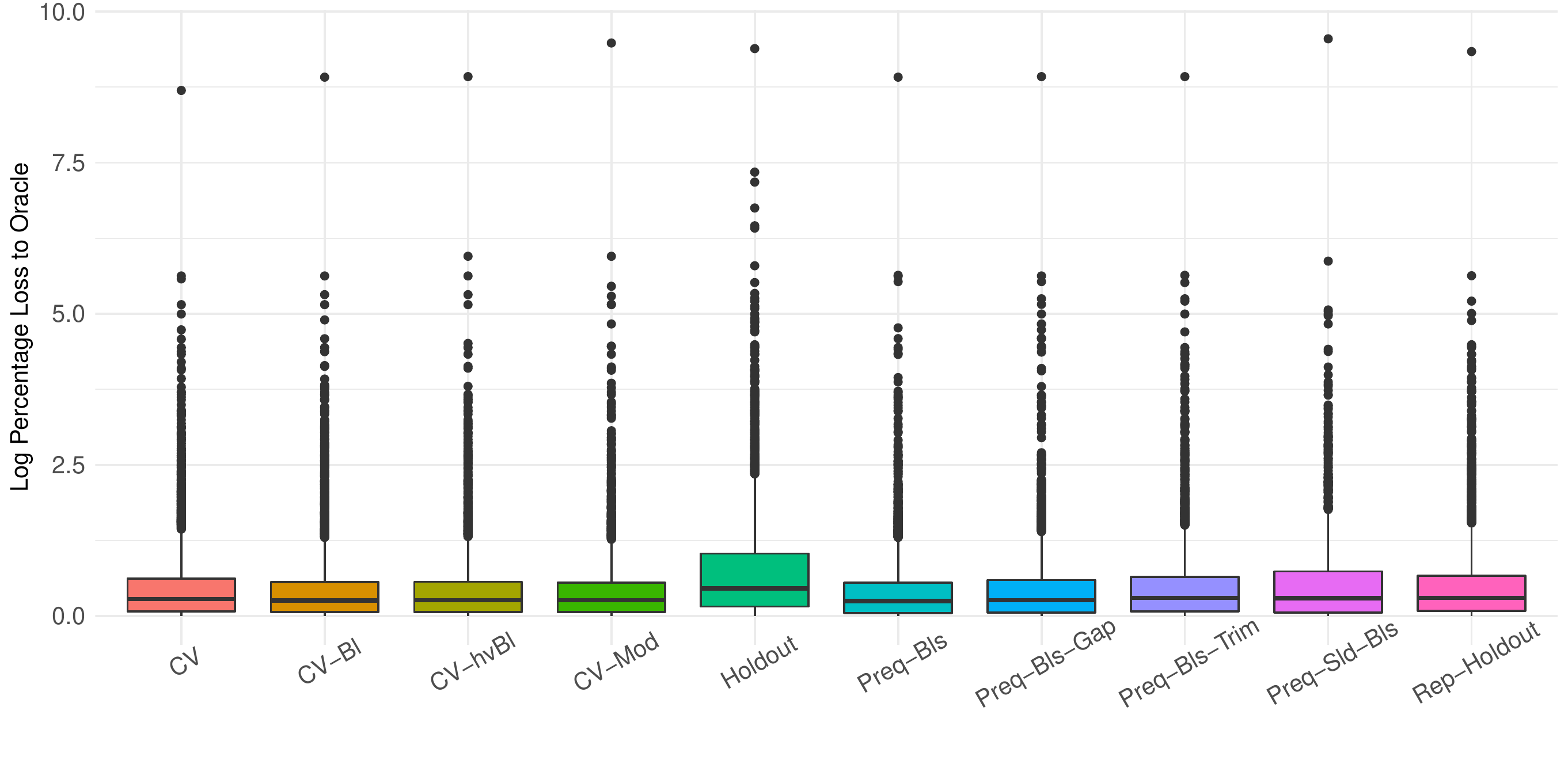}
    \caption{Distribution of the OAL (log-scaled) incurred by each estimation method with respect to the Oracle using all time series. The results suggest a large number of outliers for all methods.}
    \label{fig:lossdistgen}
\end{figure}

In general, all estimators present a considerable variability in both AL and OAL. We explore the dispersion further in Figure \ref{fig:lossdistgen}, which shows the distribution of the loss of each estimator relative to the oracle. This loss is non-negative and, as described before, it is zero when the estimator selects the best possible solution. The figure shows that all estimators incur in large errors in some data sets. This means that, while the median errors are comparable, one may be exposed to a significant performance loss irrespective of the estimation method used.

\subsubsection{RQ3: Sensitivity Analysis on the Sample Size}

In this section, we analyse the effect of the time series sample size in the experimental results. A previous study by Cerqueira et al. \cite{cerqueira2019machine} showed that the training sample size of time series is an important factor in the relative forecasting performance among different predictive models. We aim at studying this effect in estimation methods when applied for model selection. To accomplish this, we split the time series into two groups: a group which consists of 371 time series (out of 3111) with less than 1000 observations; and another group with 2740 time series with sample size above 1000 data points. Then, we carry out the previous analysis for each group of time series.

\begin{table}[ht]
\centering
\caption{Results for each method in 371 time series with sample size below 1000 observations. Methods are ordered by increasing values of overall average loss. Each metrics is bold according to which approach (average rank or average error) provides the best result for each estimator.} 
\label{tab:sample_size_below_1k}
\begin{tabular}{lrrr | rrr}
  \hline \\[0.005cm]
  & \multicolumn{3}{c}{\textbf{Average Error}} & \multicolumn{3}{c}{\textbf{Average Rank}}\\[0.1cm]
  \hline
 & \textbf{Acc.} & \textbf{AL} & \textbf{OAL} & \textbf{Acc.} & \textbf{AL} &  \textbf{OAL} \\
  \hline
\texttt{CV-hvBl} & \textbf{0.09} & 1.53$\pm$2.96 & 1.28$\pm$2.89 & 0.06 & \textbf{1.35$\pm$2.67} & \textbf{1.19$\pm$2.41} \\ 
  \texttt{CV} & 0.08 & \textbf{1.45$\pm$3.08} & 1.30$\pm$2.78 & \textbf{0.09} & 1.45$\pm$3.10 & \textbf{1.21$\pm$2.79} \\ 
  \texttt{Preq-Bls-Trim} & \textbf{0.08} & 1.66$\pm$3.29 & 1.39$\pm$3.19 & 0.07 & \textbf{1.46$\pm$2.91} & \textbf{1.30$\pm$2.76} \\ 
  \texttt{CV-Mod} & \textbf{0.07} & \textbf{1.54$\pm$2.93} & \textbf{1.33$\pm$2.95} & 0.05 & 1.54$\pm$3.14 & 1.34$\pm$3.04 \\ 
  \texttt{CV-Bl} & \textbf{0.08} & 1.55$\pm$2.97 & \textbf{1.34$\pm$2.88} & 0.06 & \textbf{1.53$\pm$2.83} & 1.38$\pm$2.70 \\ 
  \texttt{Rep-Holdout} & \textbf{0.09} & 1.64$\pm$3.22 & 1.40$\pm$2.96 & 0.07 & \textbf{1.57$\pm$2.80} & \textbf{1.36$\pm$2.74} \\ 
  \texttt{Preq-Bls-Gap} & \textbf{0.07} & 1.65$\pm$3.21 & 1.48$\pm$3.01 & 0.06 & \textbf{1.53$\pm$3.21} & \textbf{1.36$\pm$3.05} \\ 
  \texttt{Preq-Bls} & \textbf{0.08} & 1.60$\pm$2.72 & \textbf{1.39$\pm$2.65} & 0.06 & \textbf{1.59$\pm$3.13} & 1.44$\pm$2.89 \\ 
  \texttt{Preq-Sld-Bls} & \textbf{0.04} & \textbf{1.77$\pm$3.23} & \textbf{1.70$\pm$3.19} & 0.02 & 1.79$\pm$2.97 & 1.76$\pm$2.98 \\ 
  \texttt{Holdout} & 0.06 & 2.43$\pm$5.75 & 2.25$\pm$5.77 & 0.06 & 2.43$\pm$5.75 & 2.25$\pm$5.77 \\ 
   \hline
\end{tabular}
\end{table}

\begin{table}[ht]
\centering
\caption{Results for each method in 2740 time series with sample size above 1000 observations. Methods are ordered by increasing values of overall average loss. Each metrics is bold according to which approach (average rank or average error) provides the best result for each estimator.} 
\label{tab:sample_size_above_1k}
\begin{tabular}{lrrr | rrr}
  \hline \\[0.005cm]
  & \multicolumn{3}{c}{\textbf{Average Error}} & \multicolumn{3}{c}{\textbf{Average Rank}}\\[0.1cm]
  \hline
 & \textbf{Acc.} & \textbf{AL} & \textbf{OAL} & \textbf{Acc.} & \textbf{AL} & \textbf{OAL} \\ 
  \hline
\texttt{Preq-Bls} & \textbf{0.09} & \textbf{0.29$\pm$0.58} & \textbf{0.24$\pm$0.57} & 0.08 & 0.30$\pm$0.56 & 0.26$\pm$0.56 \\ 
  \texttt{CV-Bl} & \textbf{0.10} & 0.30$\pm$0.56 & 0.25$\pm$0.56 & 0.09 & \textbf{0.28$\pm$0.52} & \textbf{0.24$\pm$0.52} \\ 
  \texttt{CV-hvBl} & \textbf{0.10} & 0.30$\pm$0.57 & 0.26$\pm$0.57 & 0.09 & \textbf{0.29$\pm$0.53} & \textbf{0.24$\pm$0.51} \\ 
  \texttt{CV-Mod} & \textbf{0.07} & \textbf{0.30$\pm$0.54} & \textbf{0.26$\pm$0.53} & 0.06 & 0.30$\pm$0.56 & 0.28$\pm$0.56 \\ 
  \texttt{Preq-Bls-Gap} & \textbf{0.09} & \textbf{0.31$\pm$0.60} & \textbf{0.26$\pm$0.61} & 0.07 & 0.32$\pm$0.57 & 0.28$\pm$0.58 \\ 
  \texttt{Preq-Sld-Bls} & 0.07 & 0.33$\pm$0.81 & 0.28$\pm$0.77 & 0.07 & \textbf{0.31$\pm$0.73} & \textbf{0.27$\pm$0.69} \\ 
  \texttt{CV} & \textbf{0.09} & \textbf{0.33$\pm$0.65} & \textbf{0.29$\pm$0.62} & 0.07 & 0.35$\pm$0.60 & 0.32$\pm$0.60 \\ 
  \texttt{Preq-Bls-Trim} & \textbf{0.09} & 0.36$\pm$0.69 & \textbf{0.30$\pm$0.68} & 0.08 & \textbf{0.35$\pm$0.61} & 0.31$\pm$0.61 \\ 
  \texttt{Rep-Holdout} & \textbf{0.08} & 0.36$\pm$0.70 & \textbf{0.30$\pm$0.68} & 0.07 & \textbf{0.35$\pm$0.63} & 0.31$\pm$0.62 \\ 
  \texttt{Holdout} & 0.07 & 0.59$\pm$1.41 & 0.51$\pm$1.33 & 0.07 & 0.59$\pm$1.41 & 0.51$\pm$1.33 \\ 
   \hline
\end{tabular}
\end{table}

The results are shown in Tables \ref{tab:sample_size_below_1k} and \ref{tab:sample_size_above_1k}, in which the first presents the results for the time series with low sample size, and the second shows the results for the group of time series with sample size above 1000 observations. There are considerable differences across the two groups. All metrics are noticeably better for larger sample sizes. This result indicates that the model selection task is easier for all estimators if more data is available, which is not surprising. 

The results for the group of larger time series (more than 1000 observations) do not vary considerably with respect to the results show in Table \ref{tab:results_general} for all time series. This is expected because most of the available time series are part of this group. Notwithstanding, the group of smaller time series (less than 1000 data points) contains 371 time series, which is a considerable amount. In terms of relative performance for this group of time series, \texttt{CV-hvBl}, \texttt{CV}, and \texttt{Preq-Bls-Trim} show the best estimation ability.

An interesting result is how much \texttt{Preq-Bls-Trim} improved relative to \texttt{Preq-Bls} in the group of small time series. As a reminder, \texttt{Preq-Bls} is a common procedure used to evaluate predictive models applied in time-dependent scenarios. For example, this strategy is implemented in the widely used \textit{scikit-learn} \cite{pedregosa2011scikit} Python library as \texttt{TimeSeriesSplits}.
Our results show that \texttt{Preq-Bls-Trim}, which discards the initial iterations of \texttt{Preq-Bls}, presents better results for smaller time series while requiring less computational resources.
A possible explanation is the following. Since the sample size is low, the initial iterations of the estimation procedure may not be representative of the complete time series, and leads to a poor model selection by \texttt{Preq-Bls}. By discarding the initial iterations we keep only those folds which are more representative of the complete time series.

\subsubsection{RQ4: Analysis of the Averaging Approaches}

Regarding the comparison between average error and average rank (also known as voting cross-validation \cite{yang2007consistency}), the results are comparable. The average error leads to a systematic, but marginal, better accuracy. Across all data sets, the AL and OAL scores are comparable. The average rank shows a slightly better AL and OAL scores when the time series comprises less than 1000 observations. 
Note that, in the case of \texttt{Holdout}, the average error and average rank are identical because this estimator relies on a single iteration. 
Since the average rank requires the extra computation of computing ranks, the average error may be a preferable approach for simplicity.

\subsubsection{RQ5: Execution Time Analysis}

Finally, we analyse the execution time of each estimator. The execution time denotes the time a given estimator takes to carry out the estimation process and produce a ranking of the predictive models under comparison. 

\begin{figure}[h]
    \centering
    \includegraphics[width=0.95\textwidth, trim=0cm 0cm 0cm 0cm, clip=TRUE]{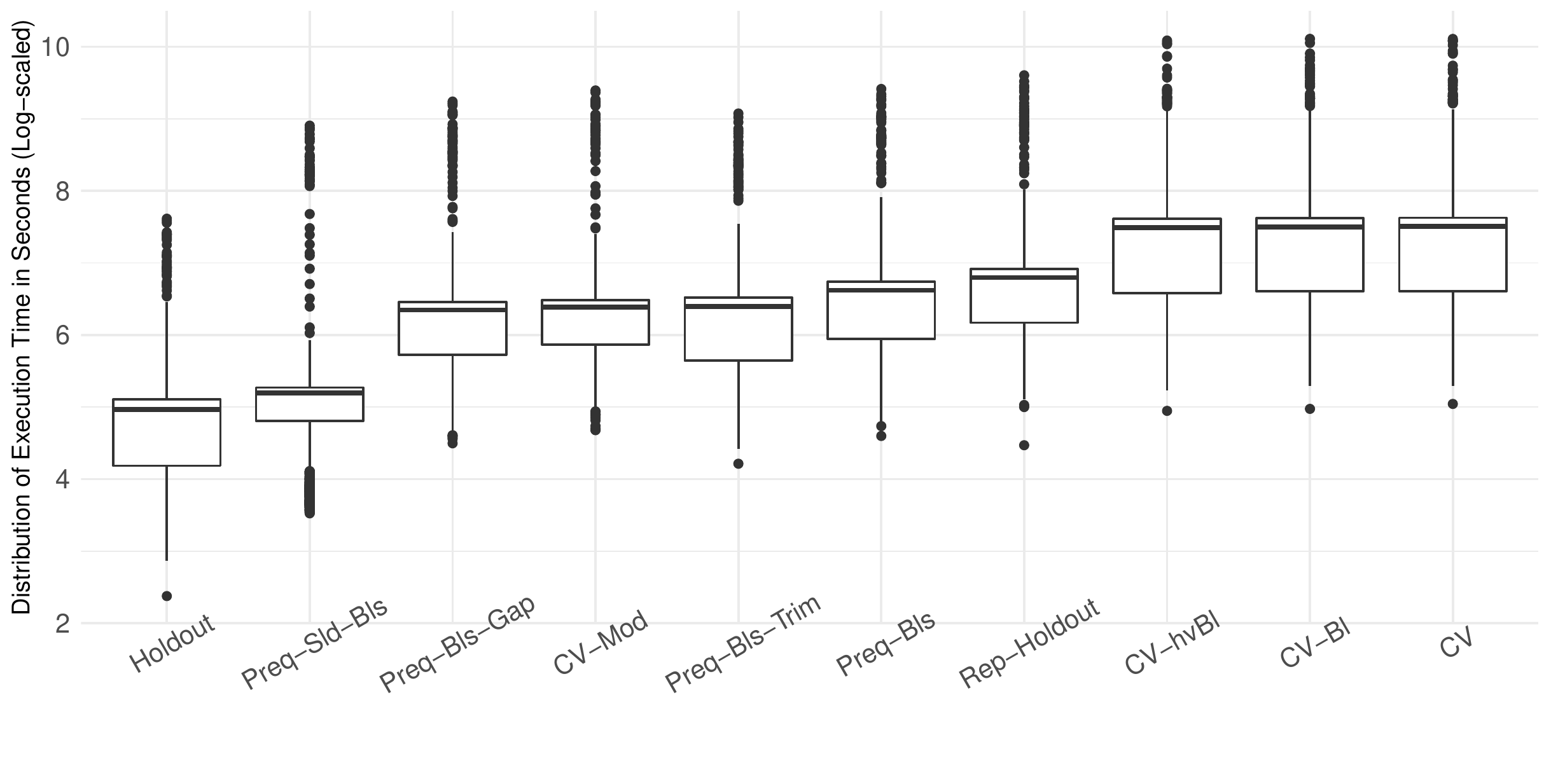}
    \caption{Distribution of the execution time in seconds (log-scaled), of each estimator across the 3111 time series.}
    \label{fig:exectime}
\end{figure}

The results of this analysis are shown in Figure \ref{fig:exectime}, which shows the distribution of the execution time of each estimator across the 3111 data sets. The estimators are ordered (from left to right) by median execution time. \texttt{Holdout} is the quickest estimator while \texttt{CV} is the slowest one, on average. In general, the execution time is correlated with the number of iterations and amount of data used by the estimator.

\section{Discussion}\label{sec:discussion}

We analysed the ability of several estimators for model selection in time series forecasting tasks. We carried out experiments using 3111 univariate time series, 10 estimation methods, and 50 auto-regressive models. We emulated a realistic scenario in which each estimator tests each one of the 50 models using the available data, and selects one for making predictions in a test set.

We discovered that, given the number of possible alternative models (50), all estimators show a low accuracy for selecting the best available model (\textbf{RQ1}) (though the scores are significantly better relative to a random selection procedure). Accordingly, this lead us to study the difference in performance between the model selected by each estimator and the model that should have been selected (the one maximizing forecasting performance on test data). Overall, the average difference across all time series ranges from about 0.28\% to about 0.58\%. However, there is a large variability in the results across all estimation methods (\textbf{RQ2}). These values may be significant in many domains, and show that the evaluation process of time series models is an important task that should not be overlooked.

We found that the results vary considerably when controlling for time series sample size. For larger time series (comprising more than 1000 observations), the results are consistent to the conclusions presented above. On the other hand, the model selection task is significantly more difficult when smaller sample sizes are available for all estimation methods. The best overall estimator incurs into an average performance loss of 0.28\%, but this value increases to 1.19\% for the group of time series with less than 1000 data points (\textbf{RQ3}).

Finally, we analysed the execution time of each approach. We found that the values are correlated with the amount of data each estimator use (\textbf{RQ4}). 



We found considerable distinctions in the relative performance of the estimators when applied for performance estimation \cite{cerqueira2020evaluating} and when used for model selection. 
For performance estimation, Cerqueira et al. \cite{cerqueira2020evaluating} report that \texttt{CV} is the worst estimator, across the 174 time series that they study. For model selection purposes, which is the topic of the current paper, \texttt{CV} is actually competitive with the best approaches, especially for smaller sample sizes.  
Cerqueira et al. \cite{cerqueira2020evaluating} did not find any impact by the time series sample size in the relative performance of the estimators. We found this factor to be important for model selection, though it is important to remark that the process for analysing this impact is different in the two studies. Moreover, this impact is noticeable both in relative terms, where the ranking of the estimators is different, and in absolute terms, in the sense that larger sample sizes lead to a better overall model selection results.

In terms of OAL there is a 0.3\% gap between the best estimator (\texttt{CV-Bl} applied with average rank) and the worst estimator (\texttt{Holdout}). This gap decreases to 0.06\% if we ignore \texttt{Holdout}, which performs systematically worse than the other approaches. 
The significance of the differences in the OAL scores diminishes further when inspecting the dispersion (using IQR) across the time series, which is often more than double the average score.
In this context, we conclude that there is no significant difference between the estimators analysed in this work. The exception is \texttt{Holdout}, which we believe should be avoided unless a large sample size is available.

\section{Conclusions}\label{sec:conclusions}

We study different estimation methods for performing model selection in time series forecasting tasks. While these methods have been studied for performance estimation \cite{bergmeir2012use,bergmeir2018note,cerqueira2020evaluating}, model selection is a different problem. Therefore, the best estimator for performing model selection may not be the most appropriate for performance estimation.

Our goal in this paper was to analyse the accuracy of each estimator for selecting the most appropriate model, and the respective forecasting performance loss when they do not. We carried out a set of experiments using 3111 time series, 10 estimation methods, and 50 predictive models to achieve this goal.

We found that the overall performance loss during model selection revolves between 0.28\% and 0.58\%. 
We also discovered that taking the average rank of models, instead of the average error, leads to a comparable performance in terms of model selection.
In terms of relative performance, we conclude that all estimation methods behave comparably for model selection as their performance differences are negligible. This means that while previous studies have show significant differences between these methods for estimating the performance of the models, when our goal is to select among them, these differences disappear. The exception is \texttt{Holdout}, which should be avoided unless there is a large sample size available.
The experiments carried out in the paper are available in an online repository.

\bibliographystyle{spmpsci}

\end{document}